\documentclass[5p,twocolumn]{elsarticle}

\usepackage{amssymb}
\usepackage{amsmath}

\usepackage{multirow}
\usepackage{varwidth}   
\usepackage{adjustbox}   
\usepackage[normalem]{ulem}
\usepackage{amsmath,amsfonts}
\usepackage{algorithmic}
\usepackage{algorithm}
\usepackage{array}
\usepackage[caption=false,font=normalsize,labelfont=sf,textfont=sf]{subfig}
\usepackage{textcomp}
\usepackage{stfloats}
\usepackage{url}
\usepackage{verbatim}
\usepackage{graphicx}
\usepackage{makecell}
\usepackage{booktabs}
\usepackage{siunitx}
\usepackage{multirow}
\usepackage[table]{xcolor} 
\journal{Knowledge-Based Systems}

\begin{document}

\begin{frontmatter}



\title{HyperEvent: A Strong Baseline for Dynamic Link Prediction via Relative Structural Encoding}


\author{Jian Gao} 
\author{Jianshe Wu} 
\author{Jingyi Ding}
\affiliation{organization={Xidian University},
            city={Xi’an},
            postcode={710071}, 
            state={Shaanxi},
            country={China}}

\begin{abstract}
Learning representations for continuous-time dynamic graphs is critical for dynamic link prediction. While recent methods have become increasingly complex, the field lacks a strong and informative baseline to reliably gauge progress. This paper proposes HyperEvent, a simple approach that captures relative structural patterns in event sequences through an intuitive encoding mechanism. As a straightforward baseline, HyperEvent leverages relative structural encoding to identify meaningful event sequences without complex parameterization. By combining these interpretable features with a lightweight transformer classifier, HyperEvent reframes link prediction as event structure recognition. Despite its simplicity, HyperEvent achieves competitive results across multiple benchmarks, often matching the performance of more complex models. This work demonstrates that effective modeling can be achieved through simple structural encoding, providing a clear reference point for evaluating future advancements. The code for HyperEvent is accessible on GitHub(\url{https://github.com/jianjianGJ/HyperEvent}).
\end{abstract}


\begin{highlights}
\item Strong and Simple Baseline. HyperEvent provides a straightforward yet competitive baseline for dynamic link prediction, using interpretable handcrafted features and a lightweight transformer. Its simplicity helps the community distinguish genuine progress from unnecessary model complexity.

\item Relative Structural Encoding. The model captures meaningful event sequences through efficient relative structural encoding—measuring direct, 1-hop, and 2-hop correlations—without relying on complex learned embeddings or memory modules.

\item Highly Scalable and Efficient. HyperEvent supports parallelized training via segmented event streams and tensorized operations, significantly reducing training time and memory usage while maintaining competitive performance across diverse temporal graph benchmarks.

\end{highlights}


\end{frontmatter}



\section{Introduction}
\label{sec_intro}

Temporal graphs model evolving systems through sequences of interactions (\textit{i.e.}, temporal events), serving as essential abstractions for real-world dynamic phenomena like social networks and transaction systems\cite{social1, social2, trade}. These graphs are primarily modeled as discrete-time graph snapshots or continuous-time dynamic graphs (CTDGs)\cite{tox}. This work focuses specifically on CTDGs because real-world interactions intrinsically manifest as continuous streams of events, where preserving their exact temporal sequence is critical for capturing fine-grained event relationships. Accurately predicting future events within CTDGs, known as dynamic link prediction, remains a core challenge with significant implications for applications such as real-time recommendation\cite{recom1,recom2} and anomaly detection\cite{anmomal1,anmomal2}.

Recent years have witnessed notable advances in representation learning for CTDGs. Proposed methods have aimed to address challenges such as capturing complex structural-temporal dependencies, modeling long-range interactions, and improving generalization to unseen nodes. These approaches can be broadly categorized into two paradigms: node-centric and event-centric methods. Node-centric methods maintain and update memory states for each node, using recurrent or attention-based mechanisms to evolve node embeddings over time\cite{tgn,jodie,dyrep,tncn}. Some incorporate neural ODEs to model continuous dynamics\cite{ctan}, or employ neighbor caching to alleviate structural bottlenecks\cite{nat}. While often scalable and inductive, these techniques reduce each event to an update of its endpoint embeddings, potentially overlooking higher-order relationships between events. In contrast, event-centric methods treat interaction events as primary entities. Techniques such as CAWN\cite{cawn} leverage causal anonymous walks to extract temporal motifs, while DyGFormer\cite{dygformer} uses attention mechanisms to implicitly model event group patterns. Several recent methods also highlight that relatively simple architectures—such as MLPs or patched transformers—can achieve strong performance\cite{graphmixer,dygformer}.

\begin{figure}[!t]
\includegraphics{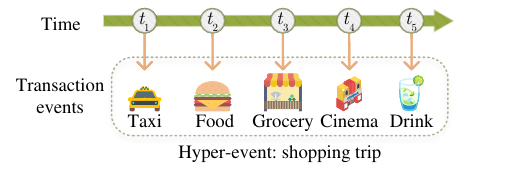}
\caption{An example of a hyper-event (shopping trip) comprising five sequential transaction events (taxi, food, grocery, cinema, drink) along a timeline ($t_1$–$t_5$).}
\label{fig_1}
\end{figure}

Among these, EdgeBank\cite{edgebank} stands out as a simple, non-parametric baseline that relies on memorizing and replaying historical interactions. However, its performance has been significantly surpassed by newer learning-based models, diminishing its utility as a meaningful benchmark. This creates a critical gap in the literature: the absence of a strong, simple baseline that can serve as a fundamental reference point for evaluating model efficacy. Without such a baseline, it becomes difficult to distinguish genuine advancements from unnecessary complexity, potentially leading to illusory progress.

An effective baseline should be conceptually clear, easy to implement, and competitive with contemporary methods. To this end, this paper introduces HyperEvent, a novel baseline model for dynamic link prediction that combines handcrafted, non-learned feature sequences with a lightweight transformer classifier. The approach is motivated by the observation that individual events often exhibit natural patterns, forming cohesive higher-order structures. For example, as illustrated in Figure~\ref{fig_1}, a sequence of transactions—such as taking a taxi, purchasing food, buying groceries, watching a movie, and having drinks—may collectively form a natural sequence of related events. HyperEvent leverages this insight by constructing feature sequences that encode relative structural patterns between the query event and relevant historical events. These features serve as input to a standard transformer that classifies whether the query event follows a plausible pattern, and hence is likely to occur.

HyperEvent offers a straightforward approach to event prediction by focusing on relative structural patterns rather than complex event relationships. This simplification allows the model to capture meaningful event sequences without relying on intricate learned embeddings or memory modules. Despite its simplicity, HyperEvent achieves competitive performance compared to state-of-the-art methods across multiple benchmark datasets. It thus fills an important gap as a strong, simple baseline that can aid the community in evaluating model improvements.

The main contributions of this work are summarized as follows:

\textbf{1) A New Simple Baseline}: HyperEvent is proposed as a straightforward baseline for dynamic link prediction that combines interpretable feature construction with a standard classifier, offering a clear reference for future research.

\textbf{2) Relative Structural Encoding}: A simple feature extraction strategy is introduced to capture relative structural patterns between the query event and historical events, enabling the model to recognize meaningful event sequences without complex parameterization.

\textbf{3) Highly Scalable Inference}: The model architecture permits efficient and parallelizable training and inference, making it suitable for large-scale temporal graphs.

\textbf{4) Extensive Empirical Validation}: Comprehensive experiments demonstrate that HyperEvent achieves competitive results across multiple datasets, often matching or exceeding more complex models.

\subsection{Related Work}
\label{subsec1}

\noindent\textbf{Temporal Graph Learning for Dynamic Link Prediction}. Dynamic link prediction is an essential task for forecasting future interactions in continuously evolving networks, with critical applications across social networks \cite{social,social3}, communication systems \cite{com1,com2}, traffic prediction \cite{traffic1,traffic2}, and knowledge graph completion \cite{knowledge1,knowledge2}. Unlike traditional static graph methods that rely on structural heuristics or node similarity metrics \cite{lp1,lp2,lp3,lp4,lp5}, dynamic settings require modeling temporal event streams, capturing complex dependencies among interactions, and generalizing to unseen nodes \cite{complex1,complex2}. Early approaches discretized event streams into snapshot sequences \cite{snap1,snap2,snap3,snap4} and applied graph neural networks \cite{gnn1,gnn2} combined with sequence models such as RNNs \cite{lstm,gru} or Transformers \cite{transfomer}. However, such discretization often leads to information loss and limited expressiveness, motivating a shift toward continuous-time modeling.

\noindent\textbf{Continuous-Time Dynamic Graph Learning}. CTDG methods can be broadly categorized into node-centric and event-centric paradigms. Node-centric approaches maintain and update memory states for each node upon every interaction. Seminal works include JODIE \cite{jodie}, which uses coupled RNNs to update user and item representations, and DyRep \cite{dyrep}, which incorporates multi-hop propagation. The Temporal Graph Network (TGN) framework unifies various memory modules and graph aggregation operators\cite{tgn}, and TGAT employs self-attention to aggregate temporal neighbor information\cite{tgat}. Though later methods such as TNCN \cite{tncn} and EDGE \cite{edge} improve scalability through asynchronous updates or graph partitioning, they remain constrained by sequential, event-by-event processing that limits parallelism and overlooks higher-order event semantics. On the other hand, event-centric methods explicitly model interactions between events. CAWN \cite{cawn} and TPNet \cite{tpnet} encode anonymous walks to capture temporal motifs. NAT introduces reusable neighbor representations\cite{nat}, and DyGFormer applies Transformers to patched event sequences\cite{dygformer}, implicitly modeling event group patterns through attention mechanisms. TCL constructs temporal dependency graphs but relies on RNNs that struggle with long-range dependencies\cite{tcl}. Neural ODE-based approaches like CTAN enable continuous dynamics modeling but often ignore fine-grained event correlations\cite{ctan}. A common trend across these lines of research is the increasing architectural complexity, with many recent models incorporating sophisticated mechanisms such as attention, memory, or neural differential equations. Despite reported improvements, the field lacks a simple, yet effective baseline that clarifies the minimal complexity required for competitive performance, making it difficult to assess whether proposed enhancements genuinely capture essential temporal-topological phenomena or merely introduce inductive biases.

\noindent\textbf{The Role of Baselines and the Limitations of EdgeBank}. Strong baselines are fundamental to meaningful scientific progress in machine learning. They serve as essential references that help the community discern actual advancements from false positives, ensure fair comparisons, and maintain a clear lower bound of performance that new methods should exceed. A poorly performing baseline can misleadingly inflate the perceived utility of new models and encourage unnecessary complexity. In temporal graph learning, EdgeBank has been adopted as a canonical baseline—a simple non-parametric method that predicts future edges based on their recurrence in historical event sequences\cite{edgebank}. While computationally efficient and easy to implement, its performance has fallen significantly behind modern learned models across multiple datasets. This growing performance gap diminishes its utility as a meaningful reference, raising concerns about pseudo-progress in the field. When a simple heuristic is substantially outperformed, it becomes unclear whether newly proposed architectures are capturing genuine temporal dynamics or simply leveraging additional parameters and complexity to memorize dataset-specific patterns. Therefore, there is a pressing need for a new, stronger, and principled baseline that is both conceptually simple and highly competitive, enabling more rigorous evaluation and providing clearer insights into what constitutes necessary model complexity for dynamic link prediction.

\section{Method}
\subsection{Preliminaries and Problem Formulation}
\noindent\textbf{Continuous-time Dynamic Graph}: A continuous-time dynamic graph represents the dynamics of interactions between entities over a continuous timeline. Formally, it is as a chronologically ordered sequence of interaction events: $\mathcal{G} = \{(u_1, v_1, t_1), (u_2, v_2, t_2), \cdots, (u_n, v_n, t_n)\}$, where $t_1 \leq t_2 \leq \cdots \leq t_n$. Each event $(u_i, v_i, t_i)$ signifies an interaction occurring at timestamp $t_i$ between a source node $u_i$ and a destination node $v_i$, both belonging to the universal node set $\mathcal{V}$. While events may possess associated features $\mathbf{e}_{u_i,v_i}^{t_i}$, our approach primarily focuses on the temporal sequence of interactions for the core problem formulation.

\noindent\textbf{Dynamic Link Prediction}: The dynamic link prediction task aims to forecast future interactions based on the observed historical evolution of the graph. Given the sequence of all events occurring strictly before a query time $t^*$, denoted as $\mathcal{E}_{<t^*} = \{ (u, v, t) \in \mathcal{G} \mid t < t^* \}$, the task is to predict the existence of a specific interaction between nodes $u^*$ and $v^*$ at time $t^*$. Formally, the goal is to determine the likelihood $P\left( (u^*, v^*, t^*) \text{ occurs} \mid \mathcal{E}_{<t^*} \right)$. This prediction should be accurate not only for interactions involving nodes observed during the historical period (transductive setting) but also for interactions potentially involving nodes unseen before $t^*$ (inductive setting).
\subsection{Existing Frameworks}
\subsubsection{Node-centric methods} \label{sec_node_centric}
Node-centric methods jointly leverage spatial and temporal models to construct dynamic node embeddings. The framework operates as follows:
\begin{equation}\label{eq1}\mathbf{h}_u(t) = \Psi_{\text{temp}} \big( \mathbf{h}_u(t^-),~ \Psi_{\text{spatial}}( \{ \mathbf{h}_v(t^-) \mid v \in \mathcal{N}_u(t) \} ) \big)
\end{equation}
where $\Psi_{\text{spatial}}$ aggregates topological information from neighboring nodes $\mathcal{N}_u(t)$ (e.g., via GNNs), $\Psi_{\text{temp}}$ updates the state using sequential models (e.g., RNNs), and $t^-$ denotes the timestamp immediately before $t$. For link prediction at $t^*$, the likelihood is computed as:
\begin{equation}P\big((u^*, v^*, t^*) \in \mathcal{E}\big) = \sigma\big( \mathbf{W}^\top [\mathbf{h}_{u^*}(t^{*-}) \oplus \mathbf{h}_{v^*}(t^{*-})] + b \big)
\end{equation}
where $\sigma$ is the sigmoid function, $\mathbf{W}$ and $b$ are learnable parameters of a linear scoring unit, $\oplus$ denotes concatenation.

\begin{figure*}[!t]
	\includegraphics{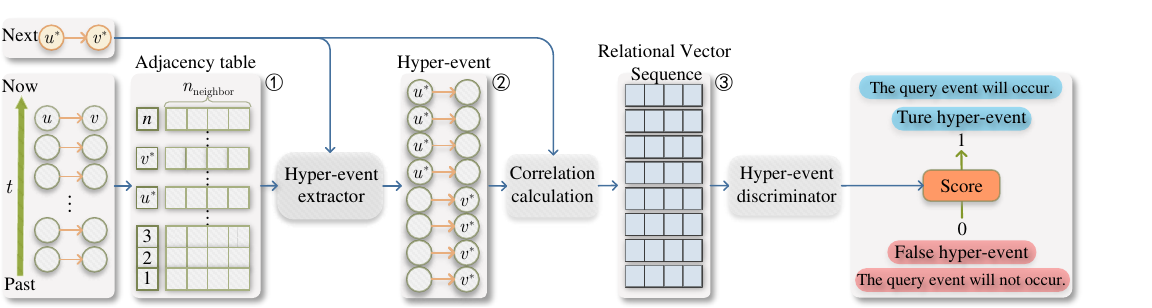}
	\caption{The HyperEvent prediction pipeline: 1)Temporal neighborhood interactions are maintained in real-time adjacency tables; 2) Hyper-events are extracted from source/destination node neighborhoods; 3)Structural-temporal correlations are computed as relational matrices; 4)A discriminator verifies hyper-event authenticity to predict query event occurrence.  }
	\label{fig_2}
\end{figure*}
\subsubsection{Event-centric Methods}
For a query event $e^* = (u^*, v^*, t^*)$, these methods first identify a relevant event subset $\mathcal{S}_{e^*}$ via neighborhood sampling:
\begin{equation}\mathcal{S}_{e^*} = \Big\{ e_k \mid \phi(u^*, v^*, u_k, v_k, |t^* - t_k|) \leq \delta, t_k < t^* \Big\},
\end{equation}
where $\phi(\cdot)$ is a heuristic proximity function (e.g., temporal or structural closeness) and $\delta$ a threshold.The event feature matrix $\mathbf{M}_{e^*} \in \mathbb{R}^{|\mathcal{S}_{e^*}| \times d}$ is constructed by aggregating features:
\begin{equation}\mathbf{M}_{e^*} = \begin{bmatrix}\mathbf{x}_1 \\\vdots \\\mathbf{x}_{|\mathcal{S}_{e^*}|}\end{bmatrix}, \quad \forall e_i \in \mathcal{S}_{e^*}.
\end{equation}
Here, $\mathbf{x}$ denotes customized event features, which can be constructed by concatenating edge features and node features, or incorporating other tailored features (such as co-neighbor encoding), along with temporal encoding.A sequence encoder (\textit{e.g.}, Transformer) processes $\mathbf{M}_{e^*}$ to produce the query's predictive representation:
\begin{equation}
	\mathbf{h}_{e^*} = \text{Encoder}(\mathbf{M}_{e^*}).
\end{equation}
The existence probability of $e^*$ is computed as:
\begin{equation}
	P\big((u^*, v^*, t^*) \in \mathcal{E}\big) = \sigma \big( \mathbf{w}^\top \mathbf{h}_{e^*} + b \big),
\end{equation}
where $\mathbf{w}$ and $b$ are learnable parameters, and $\sigma(\cdot)$ is the sigmoid function.

\textbf{The Risk of Pseudo-Progress}: While these methods have contributed significantly to the field, they often introduce substantial complexity through multi-component architectures involving both spatial and temporal modules. This complexity can obscure whether performance improvements arise from capturing genuine temporal-topological phenomena or simply from increased model capacity. Such drawbacks highlight a risk of pseudo-progress, where increasingly intricate models are proposed without a clear baseline to confirm that the added complexity corresponds to fundamental gains in understanding or capability.
\subsection{HyperEvent––A Simple Structural Encoding Baseline}
In response to the need for a straightforward yet effective baseline in dynamic link prediction, we propose a method based on relative structural encoding. This approach focuses on capturing the structural relationships between a query event and its historical context through carefully designed, handcrafted features.

\textbf{Intuition for Relative Structural Encoding}: Our approach is motivated by the observation that interactions between nodes often follow patterns that can be identified by examining their relative positions within the historical interaction graph. Specifically, we consider the following intuitive notions of structural relationships:1) \textit{Direct co-occurrence}: How frequently nodes directly interact with the query event's nodes (0-hop).2) \textit{Immediate neighborhood overlap}: How the neighborhoods of query event's nodes overlap with historical events (1-hop).3) \textit{Broader neighborhood patterns}: How the neighborhoods of neighborhoods overlap, providing a wider context (2-hop).These relative relationships can be computed efficiently and provide a rich representation of the structural context surrounding the query event, which surprisingly proves effective for dynamic link prediction.

Instead of learning stacked isolated event embeddings, HyperEvent predicts whether a query event $e^* = (u^*, v^*, t^*)$ forms a consistent \textit{hyper-event} $\mathcal{H}_{e^*}$ with relevant historical events. A hyper-event represents an integrated interaction pattern where constituent events exhibit intrinsic relational dependencies. The existence probability of $e^*$ is derived from the authenticity of $\mathcal{H}_{e^*}$:

\begin{equation}
	P\big((u^*, v^*, t^*) \in \mathcal{E}\big) = D(\mathcal{H}_{e^*})
\end{equation}
where $D(\cdot) \in \{0,1\}$ is the hyper-event discriminator.

The design philosophy of HyperEvent prioritizes conceptual clarity and computational efficiency over architectural complexity. Its pipeline, depicted in Figure \ref{fig_2}, relies on non-learned, handcrafted feature computations followed by a standard transformer classifier, eschewing the complex custom modules often found in recent works. This deliberate simplicity is a core tenet of its utility as a baseline. 

The pipeline consists of four components:1) \textit{Real-time Adjacency Table Maintenance}: Dynamically captures and updates recent neighborhood interactions to preserve the evolving graph structure. 2) \textit{Hyper-Event Extraction}: Identifies composite event patterns by jointly analyzing neighborhoods of source and destination nodes. 3) \textit{Pairwise Relational Encoding}: Generates matrices that explicitly model the structural-temporal dependencies between events using the proposed relational vectors. 4) \textit{Authenticity Discrimination}: Employs a Transformer-based architecture to evaluate the likelihood that an extracted hyper-event represents a legitimate interaction pattern.  
Subsequent sections detail the design and implementation of each component.
\subsubsection{Real-time Adjacency Table Maintenance ($\textcircled{1}$ in Figure \ref{fig_2})}\label{sec_adj} 
For efficiency and real-time processing—key attributes of a practical baseline—HyperEvent maintains a dynamic adjacency table $\mathcal{A}$. For each node $v \in \mathcal{V}$, $\mathcal{A}_v$ stores the $n_{\text{neighbor}}$ most recent interaction partners ordered chronologically:
\begin{equation}
	\mathcal{A}_v(t) = \big[ (v_1, \tau_1), \dots, (v_k, \tau_k) \big],
\end{equation}
where $t > \tau_1 > \tau_2 > \cdots > \tau_k$ and  $k \leq n_{\text{neighbor}}$.Here, $n_{\text{neighbor}}$ controls the storage overhead of the adjacency table and critically defines the historical depth considered when constructing the context for a query event, enabling efficient aggregation of relevant events. Specifically, a smaller $n_{\text{neighbor}}$ value prioritizes short-term historical interactions, enhancing sensitivity to immediate dynamics, whereas a larger value incorporates long-term contextual patterns, capturing enduring relationships. Practical event prediction scenarios exhibit distinct preferences: for instance, in highly dynamic environments like social networks where interactions evolve rapidly, a lower $n_{\text{neighbor}}$ (\textit{e.g.}, near 10) is preferable to adapt swiftly to recent changes; conversely, in stable relational settings such as collaboration networks, a higher $n_{\text{neighbor}}$ (\textit{e.g.}, near 50) better utilizes extensive historical data to improve accuracy. Empirical verification on practical applications shows that $n_{\text{neighbor}}=$10-50 suffices for most datasets, balancing computational efficiency and contextual richness.

The adjacency table enables $\mathcal{O}(1)$ event retrieval, eliminating costly neighborhood searches. Moreover, the adjacency table can be rapidly obtained since its update consists of $\mathcal{O}(1)$ element addition operations, without requiring any computations, serving as a crucial foundation for the Efficient Training Algorithm introduced in Section \ref{sec_train}.

All variables in the method are computed in real time, the time term $t$  is omitted in subsequent descriptions for conciseness.
\subsubsection{Context Extraction ($\textcircled{2}$ in Figure \ref{fig_2})} For query event $e^*=(u^*, v^*, t^*)$, extract $n_{\text{latest}}$ relevant events for both $u^*$ and $v^*$ directly from $\mathcal{A}$ :
\begin{equation}\label{eq_top_prefer}\mathcal{S}_{u^*} = \text{Top-$n_{\text{latest}}$ events from } \mathcal{A}_{u^*},
\end{equation}
\begin{equation}\mathcal{S}_{v^*} = \text{Top-$n_{\text{latest}}$ events from } \mathcal{A}_{v^*}.
\end{equation}
The context for the query event is constructed as:
\begin{equation}
	\mathcal{H}_{e^*} = \mathcal{S}_{u^*} \cup \mathcal{S}_{v^*}.
\end{equation}

\begin{figure}[!t]
	\includegraphics{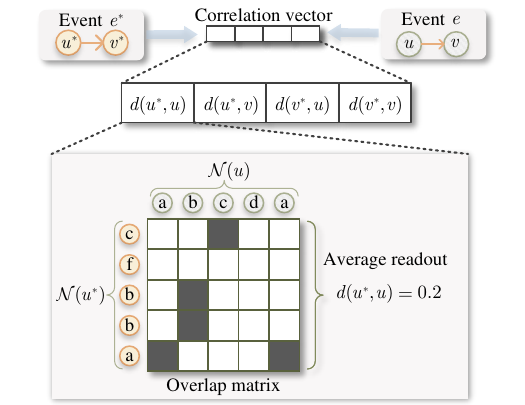}
	\caption{Visualization of an example of the event correlation vector computation in the HyperEvent.}
	\label{fig_3}
\end{figure}
\subsubsection{Relative Structural Encoding ($\textcircled{3}$ in Figure \ref{fig_2})}
For each event $e = (u, v, t) \in \mathcal{H}_{e^*}$, we compute a relative structural feature vector that captures the relationships between the historical event and the query event $e^* = (u^*, v^*, t^*)$. This vector is designed based on the intuitive notions of relative neighborhood relationships we identified earlier.

\noindent\textbf{0-hop Correlation}: This measures the direct co-occurrence between nodes. It quantifies how often a node appears in the historical interactions of another node:
\begin{equation}\label{dab_0}d^{(0)}(a,b) = \frac{1}{|\mathcal{A}_b|}  \sum_{j=1}^{|\mathcal{A}_b|} \mathbb{I}\big[ a = \mathcal{A}_b(j) \big].
\end{equation}
This provides a measure of how directly the nodes involved in historical events are related to the query event's nodes.

\noindent\textbf{1-hop Correlation}: This measures the overlap in the immediate neighborhoods of nodes:
\begin{equation}\label{dab}d^{(1)}(a,b) = \frac{1}{|\mathcal{A}_a| \cdot |\mathcal{A}_b|} \sum_{i=1}^{|\mathcal{A}_a|} \sum_{j=1}^{|\mathcal{A}_b|} \mathbb{I}\big[ \mathcal{A}_a(i) = \mathcal{A}_b(j) \big],\end{equation}
where $\mathbb{I}$ is the indicator function. This captures how the neighborhoods of the query event's nodes overlap with historical events.

\noindent\textbf{2-hop Correlation}: This extends the neighborhood analysis to include the neighbors of neighbors, providing a broader context:
\begin{equation}\label{dab_2}d^{(2)}(a,b) = \frac{1}{|\tilde{\mathcal{A}}^{(2)}_a| \cdot |\tilde{\mathcal{A}}^{(2)}_b|} \sum_{i=1}^{|\tilde{\mathcal{A}}^{(2)}_a|} \sum_{j=1}^{|\tilde{\mathcal{A}}^{(2)}_b|} \mathbb{I}\big[ \tilde{\mathcal{A}}^{(2)}_a(i) = \tilde{\mathcal{A}}^{(2)}_b(j) \big],\end{equation}
with $\tilde{\mathcal{A}}^{(2)}_a = \bigcup_{x \in \tilde{\mathcal{A}}_a} \tilde{\mathcal{A}}_x$ and $\tilde{\mathcal{A}}_a$ being the top-$\lfloor \sqrt{n_{\text{neighbor}}} \rfloor$ neighbors of $a$. This captures broader structural patterns beyond immediate neighborhoods.

\noindent\textbf{Feature Vector Construction}: The relative structural feature vector for each historical event is constructed by concatenating these three correlation measures for all relevant node pairs:
\begin{equation}\mathbf{r} =\begin{aligned}[&d^{(0)}(u^*,u),\  d^{(0)}(u^*,v),\  d^{(0)}(v^*,u),\  d^{(0)}(v^*,v), \\&d^{(1)}(u^*,u),\  d^{(1)}(u^*,v),\  d^{(1)}(v^*,u),\  d^{(1)}(v^*,v), \\&d^{(2)}(u^*,u),\  d^{(2)}(u^*,v),\  d^{(2)}(v^*,u),\  d^{(2)}(v^*,v)]\end{aligned}
\end{equation}
This 12-dimensional vector captures the relative structural relationships between the historical event and the query event. The entire context for the query event is represented as a sequence of these vectors $\mathbf{R}_{\mathcal{H}_{e^*}} = [ \mathbf{r}_1, \mathbf{r}_2, \dots, \mathbf{r}_{|\mathcal{H}_{e^*}|} ]$, preserving the chronological order of events.
\subsubsection{Classifier}To complete the baseline model, we employ a standard Transformer architecture as the classifier. This choice reflects our goal of using a powerful yet generic model, ensuring that the model's performance is driven by the quality of the input features rather than specialized architectural innovations. The Transformer processes $\mathbf{R}_{\mathcal{H}_{e^*}}$ to predict whether the query event is likely to occur:
\begin{equation}\label{eq_trans}D(\mathcal{H}_{e^*}) \in \mathcal{E}\big) = \sigma \big( \mathbf{w}^\top \text{Transformer}(\mathbf{R}_{\mathcal{H}_{e^*}}) + b \big),\end{equation}
where $\sigma$ denotes the sigmoid function. Binary cross-entropy loss is employed to optimize the classifier:
\begin{equation}\label{eq_loss}\mathcal{L} = -\sum_{\mathcal{H}_{e^*}} \Big[ y \log D(\mathcal{H}_{e^*}) + (1 - y) \log \big(1 - D(\mathcal{H}_{e^*})\big) \Big]\end{equation}
where $y = 1$ if the query event represents a true interaction, and $y = 0$ otherwise.

\begin{figure}[!t]
	\includegraphics{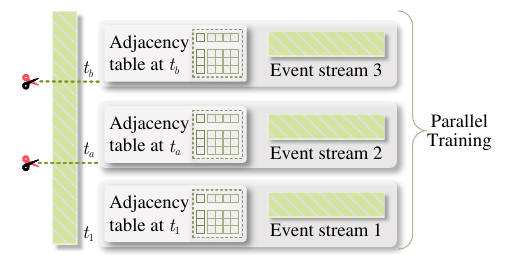}
	\caption{Parallelized training framework. Adjacency tables precomputed at key timestamps ($t_{1}, t_{2}, t_{3}$) anchor segmented event streams (Event Stream 1, Event Stream 2, Event Stream 3). }
	\label{fig_4}
\end{figure}
\subsection{Efficient Training Algorithm}\label{sec_train}The practical utility of a baseline model also depends on its trainability on large-scale datasets. To ensure our method meets this requirement, we propose an efficient training algorithm that addresses the computational challenges of continuous temporal graphs.

The proposed method fundamentally differs from existing methods through its exclusive dependence on adjacency tables rather than learned node states or historical event embeddings. All critical computations—from context extraction to prediction—operate directly on adjacency tables without maintaining incremental node-specific historical information. This design principle enables segmentation-based training: given an adjacency table at any arbitrary starting timestamp, our method can commence training from that precise temporal point without loss of continuity. Consequently, precomputed adjacency tables are only required at segment boundaries, which are efficiently obtainable through cumulative aggregation.

As illustrated in Figure \ref{fig_4}, the proposed method partitions the original event stream $\mathcal{G}$ into multiple non-overlapping segments (e.g., Event Stream 1, Event Stream 2 and Event Stream 3). Each segment corresponds to a distinct time interval demarcated by split points ($t_{1}$, $t_a$, $t_b$). Crucially, adjacency tables are precomputed at the starting timestamp of every segment, capturing the cumulative graph state up to that point. This precomputation mechanism, detailed in Section \ref{sec_adj}, ensures that the cumulative structural information captured at segmentation boundaries is functionally equivalent to that obtained during continuous sequential processing.

The adjacency tables serve as compressed representations of historical interactions preceding each segment. Our method enables fully parallelized training across all segments by decoupling segment computations through precomputed tables.

The number of segments $n_{\text{segment}}$ is a pivotal hyperparameter governing the granularity of the partitioning scheme. A larger $n_{\text{segment}}$ yields finer segmentation, reducing the average segment length at the cost of increased storage overhead for maintaining a larger number of precomputed adjacency tables. Conversely, a smaller $n_{\text{segment}}$ minimizes storage requirements but places more update steps demands on individual segments. This establishes an explicit trade-off between computational time efficiency and spatial overhead: $n_{\text{segment}}$ allows practitioners to tailor the segmentation strategy to specific hardware constraints. In practice, $n_{\text{segment}}$ is typically determined via empirical tuning to balance these opposing resource considerations while ensuring that segment boundaries do not disrupt critical temporal dependencies. The linear growth in adjacency table storage relative to $n_{\text{segment}}$ underscores the need for careful calibration of this parameter.

\noindent\textbf{Tensorized Implementation}. The parallel training paradigm fundamentally leverages tensorization rather than conventional multi-threading implementation. Crucially, all core computations within the framework operate directly on tensors, inherently enabling batched parallel execution on hardware accelerators like GPUs. Specifically, the original event stream tensor, shaped as ($|\mathcal{E}|$, 2), is partitioned into contiguous segments, transforming it into a ($n_{\text{segment}}$, $\lceil|\mathcal{E}|/n_{\text{segment}}\rceil$, 2) tensor, where padding with a designated value (-1) is applied to handle event counts not perfectly divisible by the segment count. Computational operations defined for our method naturally extend by introducing an additional segment dimension; no algorithmic redesign is required. This tensor-centric parallelism bypasses the computational overhead intrinsic to CPU multi-threading and eliminates constraints imposed by finite CPU core counts.

\begin{table*}[!t]
	\centering
	\caption{TGB Dataset Statistics}
	\label{tab:dataset}
	\begin{tabular}{lcccccc}
		\hline
		\textbf{Dataset} & \textbf{Domain} & \textbf{Nodes} & \textbf{Edges} & \textbf{Steps} & \textbf{Surprise} & \textbf{Edge Properties} \\ 
		\hline
		tgbl-wiki        & interact        & 9,227         & 157,474        & 152,757       & 0.108            & W: $\times$, Di: $\checkmark$, A: $\checkmark$ \\
		tgbl-review      & rating         & 352,637       & 4,873,540      & 6,865         & 0.987            & W: $\checkmark$, Di: $\checkmark$, A: $\times$ \\ 
		tgbl-coin        & transact       & 638,486       & 22,809,486     & 1,295,720     & 0.120            & W: $\checkmark$, Di: $\checkmark$, A: $\times$ \\ 
		tgbl-comment     & social         & 994,790       & 44,314,507     & 30,998,030    & 0.823            & W: $\checkmark$, Di: $\checkmark$, A: $\checkmark$ \\ 
		tgbl-flight      & traffic        & 18,143        & 67,169,570     & 1,385         & 0.024            & W: $\times$, Di: $\checkmark$, A: $\checkmark$ \\ 
		\hline
	\end{tabular}
\end{table*}

\subsection{Complexity Analysis}This section provides a detailed analysis of the computational efficiency of our method, examining both time and space complexity characteristics. This analysis confirms that our method achieves a highly favorable complexity profile, a necessary trait for a scalable baseline. Its efficiency stands in contrast to many more complex learning models, while its performance, as shown in Section \ref{sec_exp}, remains highly competitive.

\textbf{Time Complexity} per query event prediction is decomposed as follows: \textit{i) Adjacency Table Maintenance}: Requires $\mathcal{O}(1)$ operations per event insertion due to fixed-length neighbor storage, enabling real-time updates.\textit{ii) Context Extraction}: Achieves $\mathcal{O}(1)$ lookup by directly accessing precomputed neighbor lists, eliminating neighborhood sampling costs.\textit{iii) Relative Structural Encoding}: Consumes $\mathcal{O}(n_{\text{neighbor}}^2)$ operations per event due to pairwise neighbor comparisons, but remains independent of overall graph size. With constrained $n_{\text{neighbor}}$ and optimized 2-hop calculation using $\lfloor \sqrt{n_{\text{neighbor}}} \rfloor$ neighbors, this reduces to practical constant time.\textit{iv) Classifier Inference}: Transformer processing exhibits $\mathcal{O}((n_{\text{latest}})^2 \cdot d_{\text{model}})$ complexity per context, where $n_{\text{latest}}$ is small and the number of model channels $d_{\text{model}}$ is fixed.

The aggregate $\mathcal{O}(1)$ neighborhood operations and $\mathcal{O}(n_{\text{latest}}^2)$ model inference demonstrate independence from global graph scale $|\mathcal{V}|$ or $|\mathcal{E}|$, enabling constant-time prediction scaling.

\textbf{Space Complexity} is dominated by: \textit{i) Adjacency Tables}: Require $\mathcal{O}(|\mathcal{V}| \cdot n_{\text{neighbor}})$ storage, linear in node count but with small constant $n_{\text{neighbor}}$ (typically $\leq50$).\textit{ii) Segment Precomputation}: Efficient training adds $\mathcal{O}(n_{\text{segment}} \cdot |\mathcal{V}| \cdot n_{\text{neighbor}})$ storage for boundary adjacency tables, trading marginal space overhead for massive parallelism.\textit{iii) Classifier Parameters}: Occupy fixed $\mathcal{O}(d_{\text{model}}^2)$ space independent of graph dynamics.

The $\mathcal{O}(|\mathcal{V}|)$ spatial scaling is substantially efficient, particularly given the multiplicative constant $n_{\text{neighbor}}$ is empirically shown to be small. When combined with the segment-level parallelism described in Section \ref{sec_train}, our method achieves unprecedented scalability: The framework efficiently processes temporal graphs of tens of millions of edges through parallel training, while maintaining constant-time prediction latency for individual queries, overcoming critical bottlenecks in streaming graph learning.

\section{experiment}\label{sec_exp}
This section details the experimental framework, including datasets, baselines, evaluation protocols, and implementation specifics, followed by systematic analyses of comparative performance, ablation studies. The source code for HyperEvent is publicly accessible on GitHub\footnote{\url{https://github.com/jianjianGJ/HyperEvent}}.

\subsection{Datasets}
HyperEvent is evaluated using the Temporal Graph Benchmark (TGB) \cite{tgb}, a comprehensive repository containing five temporal graph datasets for dynamic link prediction. Following TGB's standardized protocol, all datasets are chronologically partitioned into training (70\%), validation (15\%), and test sets (15\%) to ensure temporally meaningful evaluation. The experiments span five large-scale datasets representing diverse real-world dynamics:
\textbf{1) Wiki} (tgbl-wiki-v2): Captures bipartite user-page interactions from Wikipedia co-editing logs, featuring text-attributed edges\cite{jodie}.
\textbf{2) Review} (tgbl-review-v2): Contains weighted user-product interactions from Amazon electronics reviews (1997–2018)\cite{rewiew}.
\textbf{3) Coin} (tgbl-coin-v2): Documents cryptocurrency transactions during the 2022 Terra Luna market crash\cite{coin}.
\textbf{4) Comment} (tgbl-comment): Models directed reply networks from Reddit (2005–2010), comprising over 44 million edges\cite{comment}.
\textbf{5) Flight} (tgbl-flight-v2): Represents global airport traffic (2019–2022) with geo-attributed nodes and flight-number edges\cite{flight}.

As detailed in Table \ref{tab:dataset}, these graphs exhibit substantial heterogeneity in scale (638K–18M nodes, 157K–67M edges), temporal dynamics (1,385–30M timestamps), and edge characteristics (weighted, directed, and attributed). The dynamic link prediction task consistently requires forecasting the next event given historical event streams.

\subsection{Compared Methods}
HyperEvent is compared against the following temporal graph representation learning and dynamic link prediction methods, as introduced in Section \ref{sec_intro}:

\noindent\textit{Non-parametric methods}: EdgeBank$_{\infty}$ , EdgeBank$_{\text{tw}}$ \cite{edgebank}.

\noindent\textit{Deep learning methods}: JODIE \cite{jodie}, DyRep \cite{dyrep}, TGAT \cite{tgat}, TGN \cite{tgn}, CAWN \cite{cawn}, NAT \cite{nat}, TCL \cite{tcl}, GraphMixer \cite{graphmixer}, DyGFormer \cite{dygformer}, TNCN \cite{tncn}, CTAN \cite{ctan}, TPNet \cite{tpnet}.

\subsection{Evaluation Metric}
The filtered Mean Reciprocal Rank (MRR) as specified by TGB \cite{tgb} is adopted, representing a ranking-based metric particularly suited for dynamic link prediction. Unlike Area Under the ROC Curve or Average Precision, which fail to adequately capture the relative ranking of positive edges against negatives, MRR directly measures the reciprocal rank of the true destination node among negative candidates. This aligns with recommendation systems and knowledge graph completion tasks where ranking quality is paramount. Crucially, all negative edges are predefined in TGB benchmark, ensuring fair comparison across methods.

Formally, for each positive query edge with prediction score $y_{\text{pos}}$, and a set of negative candidate scores $y_{\text{neg}}$, HyperEvent computes
\textit{Optimistic rank}: Number of negatives scoring strictly higher than $y_{\text{pos}}$.
\textit{Pessimistic rank}: Number of negatives scoring at least $y_{\text{pos}}$.

The final rank $r$ is averaged:
\begin{equation}
	r = \frac{1}{2} (\text{optimistic\_rank} + \text{pessimistic\_rank}) + 1.
\end{equation}

MRR is then calculated as:
\begin{equation}
	\text{MRR} = \frac{1}{|Q|} \sum_{i=1}^{|Q|} \frac{1}{r_i},
\end{equation}
where $Q$ is the set of test queries. MRR ranges within (0,1], with higher values indicating superior ranking performance. This metric reflects the growing consensus in recent graph representation learning literature for assessing link prediction quality.

\subsection{Experimental Setup}
\begin{table}[t]
	\centering
	\small
	\setlength{\tabcolsep}{3pt} 
	\caption{HyperEvent Parameter Configurations on Different Datasets}
	\label{tab:params}
	\begin{tabular}{lccccc}
		\toprule
		\multirow{2}{*}{Parameter} & \multicolumn{5}{c}{Dataset} \\
		\cmidrule(lr){2-6}
		& Wiki & Review & Coin & Comment & Flight \\
		\midrule
		$n_{\text{neighbor}}$ & 15 & 10 & 40 & 10 & 50 \\
		$n_{\text{latest}}$ & 10 & 10 & 10 & 10 & 10 \\
		$n_{\text{segment}}$ in Train & 50 & 50 & 50 & 20 & 50 \\
		$n_{\text{segment}}$ in Val/Test & 1 & 10 & 20 & 10 & 20 \\
		\bottomrule
	\end{tabular}
\end{table}

\noindent\textbf{Dataset Configuration}. All experiments implement the standardized evaluation protocols from TGB \cite{tgb}. This includes chronological train/validation/test splits and standardized negative queries for dynamic link prediction. While several datasets include node features or edge attributes/weights, such information is deliberately excluded. This choice isolates and highlights HyperEvent's capacity to capture intrinsic event interdependencies, demonstrating the framework's efficacy without relying on external features.

\noindent\textbf{HyperEvent Parameters}. HyperEvent employs three key hyperparameters: $n_{\text{neighbor}}$, $n_{\text{latest}}$, and $n_{\text{segment}}$. The hyperparameter $n_{\text{segment}}$ primarily governs computational efficiency by segmenting the event stream for parallel processing. Selection of $n_{\text{segment}}$ balances GPU memory constraints with training duration relative to dataset scale. During evaluation, which involves processing multiple negative samples concurrently, a smaller $n_{\text{segment}}$ value is typically used compared to training. For the Wiki, Review, and Comment datasets, the optimal $n_{\text{neighbor}}$ was searched within \{10, 15, 20\}, with $n_{\text{latest}}$ fixed at 10. Observations indicated a preference for larger neighborhoods on Coin and Flight datasets, prompting a search over \{10, 15, 20, 30, 40, 50\} for $n_{\text{neighbor}}$. The hyperparameter $n_{\text{latest}}$ was optimized from \{5, 8, 10\} when fixing $n_{\text{neighbor}} = 10$. Detailed hyperparameter configurations for all datasets are provided in Table \ref{tab:params}.

\noindent\textbf{Transformer Architecture}. The hyper-event discriminator, implemented via Eq. (\ref{eq_trans}), utilizes a Transformer module. As this component does not represent a core innovation, a fixed architectural configuration was employed across all datasets without dataset-specific tuning. The settings include: a maximum sequence length of 512 for position encoding, position encoding dimension of 64, Transformer hidden dimension of 64, 3 layers, 4 attention heads, dropout rate of 0.1, and GELU activation functions.

\noindent\textbf{Training Protocol}. Optimization employed Adam with a fixed learning rate of 0.0001. The model was trained for 50 epochs on the Wiki dataset and 5 epochs on all other datasets. Model selection was based on performance on the validation set, with final results reported on the test set using the MRR metric. The mean and standard deviation of MRR over 5 independent runs with different random seeds are reported to ensure statistical reliability.

\noindent\textbf{Hardware}. All experiments were conducted on a high-performance computing system equipped with an Intel Core i7-14700KF CPU, 64 GB RAM, and an NVIDIA GeForce RTX 4090 GPU with 24 GB VRAM.

\noindent\textbf{MRR($\mathbf{\times 100}$)}. For clearer visualization in tables and figures, all presented MRR values are scaled by a factor of 100; textual references to MRR subsequently implicitly denote these scaled values.

\begin{table}[!t]
	\caption{MRR of methods on different datasets\label{tab:main}}
	\setlength{\tabcolsep}{3pt}
	\centering
	\footnotesize
	\setlength{\tabcolsep}{1pt} 
	\begin{tabular}{lccccc}
		\hline
		Model         			& Wiki & Review & Coin & Comment & Flight \\ \hline
		EdgeBank$_{\infty}$    	&49.5           & 2.3           &45.2           &12.9	  		&16.7        \\
		EdgeBank$_{\text{tw}}$ 	&57.1           & 2.5           &35.9           &14.9     		&38.7         \\
		JODIE         			&63.1$\pm$1.7	& 34.7$\pm$0.1  &OOM            &OOM      		&OOM       \\
		DyRep         			&5.0$\pm$1.7  	& 22.0$\pm$3.0	&45.2$\pm$4.6   &28.9$\pm$3.3   &55.6$\pm$1.4        \\
		TGAT         			&14.1$\pm$0.7	& 35.5$\pm$1.2	&OOM    		&OOM     		&OOM       \\
		TGN           			&39.6$\pm$6.0	& 34.9$\pm$2.0	&58.6$\pm$3.7   &37.9$\pm$2.1   &70.5$\pm$2.0        \\
		CAWN          			&71.1$\pm$0.6	& 19.3$\pm$0.1	&OOM            &OOM            &OOM       \\
		NAT           			&74.9$\pm$1.0	& 34.1$\pm$2.0	&OOM            &OOM            &OOM       \\
		TCL           			&20.7$\pm$2.5	& 19.3$\pm$0.9	&OOM            &OOM            &OOM       \\
		GraphMixer    			&11.8$\pm$0.2	& \textbf{52.1}$\pm$\textbf{1.5}	&OOM            &OOM            &OOM        \\
		DyGFormer     			&79.8$\pm$0.4	& 22.4$\pm$1.5	&75.2$\pm$0.4   &67.0$\pm$0.1   &OOM        \\
		TNCN          			&71.8$\pm$0.4	& 37.7$\pm$1.0	&76.2$\pm$0.4   &69.7$\pm$0.6  &82.0$\pm$0.4        \\
		CTAN          			&66.8$\pm$0.7	& \underline{40.5$\pm$0.4}	&74.8$\pm$0.4   &67.1$\pm$6.7   &OOM        \\ \hline
		HyperEvent(ours)    			&\underline{80.9$\pm$0.1}		&26.8$\pm$0.4			&\underline{77.3$\pm$0.3}			&\underline{75.9$\pm$0.2}			&\underline{87.7$\pm$0.3}        \\ 
		TPNet(SOTA)          			&\textbf{82.7}$\pm$\textbf{0.1}	& 35.6$\pm$0.5	&\textbf{83.2}$\pm$\textbf{0.1}   &\textbf{82.5}$\pm$\textbf{0.6}   &\textbf{88.4}$\pm$\textbf{0.1}	      \\ \hline
	\end{tabular}
\end{table}

\subsection{Methods Comparison}
Table \ref{tab:main} presents the MRR of HyperEvent against 13 methods. HyperEvent demonstrates strong performance, achieving competitive results on four datasets: Wiki (80.9), Coin (77.3), Comment (75.9), and Flight (87.7), where it is outperformed only by the TPNet model. This performance is achieved through a simple framework that leverages relative structural encoding features and a lightweight transformer classifier. The fact that this straightforward approach achieves competitive results highlights the effectiveness of relative structural encoding in dynamic link prediction.

However, on the Review dataset, HyperEvent achieves an MRR of 26.8, underperforming methods like GraphMixer (52.1) and CTAN (40.5). This phenomenon can be attributed to two dataset-specific factors:

\noindent\textit{1) High Temporal Surprise}: With a surprise metric of 0.987 (the highest among datasets), Review exhibits highly unpredictable interactions. HyperEvent's reliance on historical event patterns becomes less effective when future events deviate substantially from past structures.

\noindent\textit{2) Sparse Event Context}: Predicting "which product a user reviews next" relies heavily on individual user behavior rather than inter-event dependencies. This misalignment with HyperEvent's strength in modeling group dynamics reduces its relative advantage.

This result is instructive for future research: it clearly delineates a scenario (high surprise, sparse context) where a simple, relative-structural-based baseline may be insufficient, and more specialized models are warranted. Thus, HyperEvent serves not only as a performance reference point but also as a diagnostic tool to identify problem characteristics that demand more sophisticated solutions.

The strong performance on Coin, Comment, and Flight—datasets featuring dense event streams and strong temporal dependencies—reinforces HyperEvent's effectiveness in scenarios where event interdependencies drive evolutionary patterns. HyperEvent establishes a new, simple, and effective baseline for temporal graph representation learning.

\subsection{Effectiveness of the Training Algorithm}
\begin{table}[t]
	\centering
	\small
	\setlength{\tabcolsep}{3pt}
	\caption{Efficiency of HyperEvent on Flight Dataset}
	\label{tab:efficent}
	\begin{tabular}{lccc}
		\toprule
		\multirow{2}{*}{Method} & \multicolumn{3}{c}{Metric} \\
		\cmidrule(lr){2-4}
		& \makecell[c]{Training Time \\per Epoch} & GPU Memory & MRR \\
		\midrule
		TNCN & 10h 34m 37s & 15,189MB & 82.0 \\
		HyperEvent(ours) & \textbf{1h 4m 33s} & \textbf{2,346MB} & \textbf{87.7} \\
		\midrule
		Improvement & 89.83\% $\downarrow$ & 84.55\% $\downarrow$ & 6.95\% $\uparrow$ \\
		\bottomrule
	\end{tabular}
\end{table}

\begin{figure}[t]
	\centering
	\includegraphics{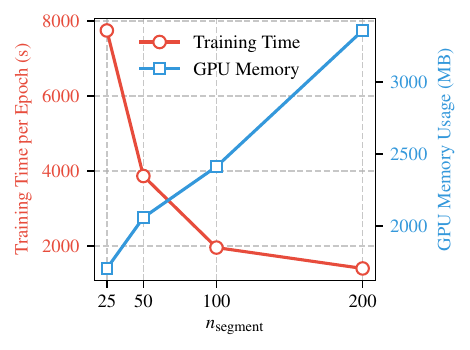}
	\caption{GPU memory usage (MB, right axis) and per-epoch training duration (seconds, left axis) of HyperEvent across varying segment counts ($n_{\text{segment}}$). }
	\label{fig_5}
\end{figure}

The efficiency gains of HyperEvent are quantified in Table \ref{tab:efficent}, contrasting HyperEvent with TNCN. HyperEvent achieves a remarkable 89.83\% reduction in per-epoch training time (1h 4m 33s vs. TNCN's 10h 34m 37s) and an 84.55\% reduction in GPU memory consumption (2,346MB vs. 15,189MB), while simultaneously improving prediction accuracy with a 6.95\% higher MRR (87.7 vs. 82.0). These results underscore the practicality of HyperEvent as a baseline. Its efficiency is not merely an ancillary benefit but a core feature, ensuring that it can be rapidly trained and evaluated on large-scale graphs, making it an accessible and scalable reference point for the research community.

The efficiency stems from three key aspects: First, segmenting the event stream into parallelizable segments reduces the iterations per epoch by a factor of $n_{\text{segment}}$. Second, a tensorized implementation leverages GPU parallelism directly, bypassing multi-threading bottlenecks and CPU-GPU communication latency. Third, the low-dimensional design of the relative structural encoding vectors (only 12 dimensions) and the lightweight classification model significantly reduce the memory footprint.

Further analysis of computational trade-offs is illustrated in figure \ref{fig_5}, which plots GPU memory usage and per-epoch training duration against varying segment counts ($n_{\text{segment}}$). The segment count critically balances memory efficiency and training speed: increasing $n_{\text{segment}}$ linearly escalates memory demands (blue curve), as $n_{\text{segment}}$ adjacency tables must be stored concurrently. Conversely, training time initially decreases nearly linearly (red curve, $n_{\text{segment}} < 100$) due to heightened parallelism. Beyond this threshold, diminishing returns emerge as GPU computational bottlenecks saturate; further increasing $n_{\text{segment}}$ yields marginal speed improvements despite continued memory growth. This underscores the algorithm's adaptability, where $n_{\text{segment}}$ can be tuned to align with hardware constraints.

Collectively, these results demonstrate that HyperEvent establishes a new benchmark for computationally efficient and effective temporal graph learning. It provides a compelling baseline that is simple in its core idea and highly efficient in practice, setting a clear reference point for future work.

\begin{figure}[!t]
	\centering
	\includegraphics{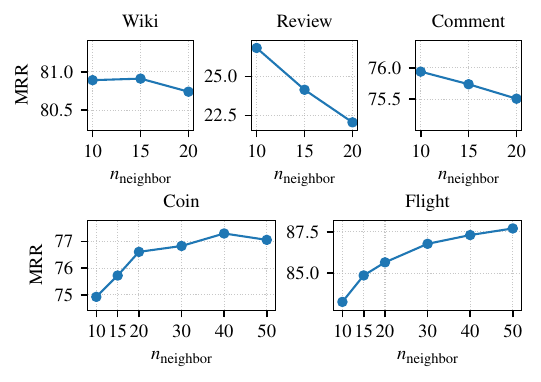}
	\caption{MRR across different datasets under varying $n_{\text{neighbor}}$ (fix $n_{\text{latest}}=10$).}
	\label{fig_6}
\end{figure}

\subsection{Modulating Long- and Short-Term Preferences via $n_{\text{neighbor}}$ and $n_{\text{latest}}$}

\subsubsection{Influence of $n_{\text{neighbor}}$} \label{sec_neighbor}
The influence of historical context depth on dynamic link prediction performance is investigated by analyzing the impact of varying the hyperparameter $n_{\text{neighbor}}$ across diverse datasets. As shown in figure \ref{fig_6}, which plots MRR under different $n_{\text{neighbor}}$ values, HyperEvent framework exhibits distinct patterns that reflect the varying preferences of datasets for long- or short-term historical information. Specifically, on the Wiki dataset, MRR demonstrates remarkable stability, with values remaining around 80.5 to 81.0 across $n_{\text{neighbor}}$ settings. Conversely, the Review dataset shows a pronounced decline, decreasing from 26.8 to 22.0 as $n_{\text{neighbor}}$ increases, indicating a sensitivity to shorter history lengths. For the Comment dataset, the change is gradual, with MRR slipping from 75.9 to 75.5. In contrast, the Coin dataset benefits from a longer context, rising steadily from 74.9 to 77.0, while the Flight dataset displays consistent improvement, climbing from 83.3 to 87.7 with higher $n_{\text{neighbor}}$.

This divergent behavior originates from the role of $n_{\text{neighbor}}$ in defining the temporal scope for constructing event sequences, as introduced in Section \ref{sec_adj}. Here, $n_{\text{neighbor}}$ represents the number of most recent interaction partners stored chronologically in the adjacency table, which directly determines the historical length considered for each node. In HyperEvent, this parameter modulates the input to the relative structural encoding calculations from equations (\ref{eq_top_prefer}) to (\ref{eq_correlation}), thereby balancing long-term preferences, such as persistent node affinities over extended timelines, against short-term preferences that focus on immediate, transient interactions. Datasets like Coin and Flight, characterized by slowly evolving patterns, achieve higher MRR with larger $n_{\text{neighbor}}$ values, as they leverage richer historical context to identify event patterns. In contrast, datasets such as Review and Comment, which involve rapid, high-frequency interactions, perform better with shorter histories, as excessive context may introduce noise that obscures recent relevant events. The findings underscore the adaptability of HyperEvent in capturing dynamic graph dynamics, highlighting how tuning $n_{\text{neighbor}}$ effectively addresses the inherent temporal heterogeneities across applications.

\begin{figure}[!t]
	\centering
	\includegraphics{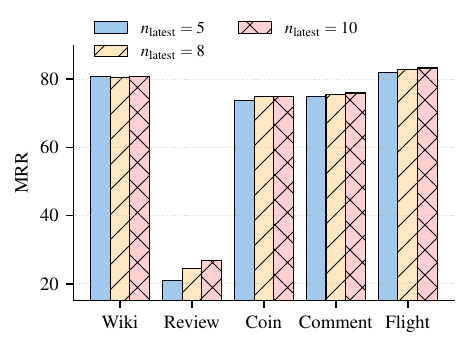}
	\caption{MRR across different datasets under varying $n_{\text{latest}}$ (fix $n_{\text{neighbor}}=10$).}
	\label{fig_7}
\end{figure}

\subsubsection{Influence of $n_{\text{latest}}$}
Figure \ref{fig_7} evaluates MRR under varying values of $n_{\text{latest}}$ (5, 8, and 10). The figure illustrates that MRR exhibits minimal variation across these settings for most datasets. Specifically, on Wiki, Coin, Comment, and Flight, the MRR values remain stable with only slight fluctuations as $n_{\text{latest}}$ increases, indicating limited sensitivity to this parameter. The sole exception is the Review dataset, where MRR shows notable differences—rising from the baseline at $n_{\text{latest}}=5$ to higher values at $n_{\text{latest}}=10$.

This behavior can be attributed to the definition of $n_{\text{latest}}$, which directly governs the number of recent events aggregated into a sequence. As elaborated in Section \ref{sec_neighbor}, datasets primarily rely on $n_{\text{neighbor}}$ to encode long- and short-term historical preferences; $n_{\text{neighbor}}$ effectively captures node-level temporal dependencies, thereby reducing the influence of $n_{\text{latest}}$ on overall predictive performance. Furthermore, excessively increasing $n_{\text{latest}}$ escalates computational overhead due to the linear scaling of event aggregation in sequence formation, without delivering commensurate gains in accuracy for most scenarios. Consequently, adopting $n_{\text{latest}}=10$ strikes an optimal balance, providing robust predictive capabilities while maintaining computational efficiency across diverse dynamic graph environments.

\subsection{Ablation Study}
\begin{table}[t]
	\centering
	\caption{MRR of HyperEvent with/without Enhanced Correlation Vector}
	\label{tab:d12}
	\begin{tabular}{lccr}
		\hline
		\multirow{2}{*}{Dataset} & \multicolumn{2}{c}{Enhanced Correlation Vector} & \multicolumn{1}{c}{\multirow{2}{*}{Drop}} \\ \cline{2-3}
		& with  & without & \multicolumn{1}{c}{} \\ \hline
		Wiki    & 80.90 & 77.62   & 4.06\%               \\
		Review  & 26.83 & 8.61    & 67.91\%              \\
		Coin    & 74.92 & 66.23   & 11.60\%              \\
		Comment & 75.94 & 53.83   & 29.12\%              \\
		Flight  & 83.26 & 74.27   & 10.80\%              \\ \hline
	\end{tabular}
\end{table}

\subsubsection{Without the Relative Structural Encoding Vector}
This ablation evaluates the contribution of the relative structural encoding vector by comparing model performance when replacing the full 12-dimensional relative structural encoding vector (integrating equations \ref{dab}, \ref{dab_0}, and \ref{dab_2}) with a reduced version using only 1-hop neighborhood overlap distance (Equation \ref{dab}). As Table \ref{tab:d12} indicates, removing multi-hop structural features causes consistent and substantial performance degradation. The MRR drops range from 4.06\% on Wiki to a drastic 67.91\% on Review, with all five datasets exhibiting declines exceeding 10\% except Wiki. This degradation proves particularly severe for datasets like Review and Comment, where node interactions involve complex higher-order dependencies.

Two key factors explain this sensitivity. First, limiting encoding to immediate (1-hop) neighborhoods overlooks critical structural contexts captured by 0-hop (node identity) and 2-hop (extended topology) features. For example, on Review, where users interact sequentially across diverse products, 0-hop identity matching helps confirm recurring user/item relationships, while 2-hop signals reveal indirect influence chains. Second, the impoverished 4-dimensional vector lacks granularity to discriminate between genuine event sequences and coincidental event aggregations. Consequently, the model frequently misclassifies valid query events that rely on multi-scale evidence, especially for inductive cases involving unseen nodes where 1-hop patterns alone are insufficient to confirm sequence integrity. These results validate that comprehensive relative structural encoding—spanning local to global neighborhood contexts—is beneficial for effective event sequence recognition.

\subsubsection{Without the Efficient Training Algorithm}
Table \ref{tab:seg_ablation} quantitatively compares per-epoch training durations for HyperEvent with and without the proposed efficient training algorithm across five temporal graph datasets. A striking observation emerges: enabling this algorithm consistently reduces training times by over 95\% across all datasets, achieving relative savings exceeding 98\% on four benchmarks. Notably, the degree of acceleration correlates with the segmentation parameter $n_{\text{segment}}$. Higher $n_{\text{segment}}$ values yield more significant efficiency gains, as evidenced by the 98.07\% saving on Flight ($n_{\text{segment}}=50$) versus 95.03\% on Comment ($n_{\text{segment}}=20$).

This dramatic improvement stems from the algorithm's core design: partitioning the monolithic event stream into $n_{\text{segment}}$ independent segments linearly reduces the effective event sequence length processed within each iteration from $|\mathcal{E}|$ to $\frac{ |\mathcal{E}|}{n_{\text{segment}}}$. Simultaneously, tensorized implementation in Section \ref{sec_train} can fully leverage the massive parallel computing capabilities of GPUs while avoiding multithreading overhead. Consequently, the computational complexity per epoch scales inversely with $n_{\text{segment}}$, explaining the near-linear time reduction observed. Such efficiency is imperative for scaling HyperEvent to massive real-world temporal graphs, where raw training times without segmentation proved prohibitive (e.g., exceeding 55 hours per epoch on Flight). This approach transforms HyperEvent from a conceptual framework into a practically deployable solution for industrial-scale dynamic graphs.

\begin{table}[t]
	\centering
	\small 
	\setlength{\tabcolsep}{1pt} 
	\caption{Per-epoch Training Time of HyperEvent with/without Efficient Training Algorithm}
	\label{tab:seg_ablation}
	\begin{tabular}{lrrr}
		\hline
		\multirow{2}{*}{Dataset} & \multicolumn{2}{c}{Efficient Training Algorithm}       & \multicolumn{1}{c}{Time Saved} \\ \cline{2-3}
		& \multicolumn{1}{c}{with} & \multicolumn{1}{c}{without} & \multicolumn{1}{c}{Relative}   \\ \hline
		Wiki    & 9s ($n_{\text{segment}}=50$)         & 7m44s      & 98.08\% \\
		Review  & 4m43s ($n_{\text{segment}}=50$)     & 4h02m19s  & 98.06\% \\
		Coin    & 22m05s ($n_{\text{segment}}=50$)    & 18h49m02s & 98.04\% \\
		Comment & 1h47m48s ($n_{\text{segment}}=20$) & 35h47m28s & 95.03\% \\
		Flight  & 1h04m45s ($n_{\text{segment}}=50$) & 55h49m05s & 98.07\% \\ \hline
	\end{tabular}
\end{table}

\section{conclusion}
In this work, we observe that many recent advances in temporal graph representation learning for dynamic link prediction are evaluated without comparison to a consistently strong and efficiently reproducible baseline. This makes it challenging to assess whether performance gains stem from genuinely useful innovations or from incidental architectural complexity. To help address this, we introduce HyperEvent, a simple and lightweight baseline that encodes relative structural patterns among recent events using handcrafted features and a minimal transformer backbone.

Rather than attempting to model complex temporal dynamics or causal relationships, HyperEvent focuses on capturing basic local structural regularities in the evolving graph through relative positional and relational encodings. Surprisingly, this straightforward approach achieves competitive or superior performance across five established TGB benchmarks, often with significantly fewer parameters and faster training than more elaborate learned models.

We do not claim that HyperEvent captures deep temporal semantics or event level dependencies. Its effectiveness suggests that simple structural priors, when carefully designed, can serve as a surprisingly strong foundation. We hope HyperEvent can act as a pragmatic reference point for future work, helping to calibrate progress, reduce redundant complexity, and encourage clearer ablation studies.

Limitations remain, particularly in settings where graph evolution is highly irregular or lacks discernible local structure. We look forward to seeing future methods that reliably surpass this baseline under such conditions. Ultimately, our goal is not to propose a final solution, but to offer a humble and reproducible starting point that helps the community build more meaningfully on solid ground.



\bibliographystyle{elsarticle-num} 
\bibliography{references}

\end{document}